\pdfoutput=1
\documentclass{article}
\PassOptionsToPackage{round}{natbib}
\usepackage[preprint]{nips_2018}

\usepackage{color}
\definecolor{darkblue}{rgb}{0,0.08,0.45}
\usepackage[colorlinks=true,allcolors=darkblue]{hyperref}       
\usepackage{url}            
\usepackage{booktabs}       
\usepackage{amsfonts}       
\usepackage{nicefrac}       
\usepackage{microtype}      

\renewcommand{\cite}{\citep}
\usepackage{amsmath}
\usepackage{amsthm}
\usepackage{newtxtext,newtxmath}
\usepackage{braket}
\usepackage{mathtools}
\usepackage[ruled,vlined,linesnumbered]{algorithm2e}
\usepackage{wrapfig}
\usepackage{siunitx}
\usepackage{tabularx}
\usepackage{etoolbox}
\newcommand{\ubold}{\fontseries{b}\selectfont}
\robustify\ubold
\newcommand{\Bf}[1]{\multicolumn{1}{l}{\ubold \num[mode=text,detect-weight,detect-inline-weight=text,tight-spacing]{#1}}}

\newtheoremstyle{mydefinition}
  {5.5pt} 
  {0pt} 
  {} 
  {} 
  {\bfseries} 
  {.} 
  {.5em} 
  {} 
\newtheoremstyle{mytheorem}
  {5.5pt} 
  {0pt} 
  {\itshape} 
  {} 
  {\bfseries} 
  {.} 
  {.5em} 
  {} 
\theoremstyle{mydefinition}

\theoremstyle{mytheorem}
\newtheorem{theorem}{Theorem}
\newtheorem{proposition}{Proposition}

\newcounter{enum2}
\renewenvironment{enumerate}{%
\begin{list}%
 {\arabic{enum2}.\ \,}{%
 \usecounter{enum2}
 \setlength{\itemindent}{0pt}
 \setlength{\leftmargin}{12pt}
 \setlength{\rightmargin}{0pt}
 \setlength{\labelsep}{0pt}
 \setlength{\labelwidth}{20pt}
 \setlength{\itemsep}{0pt}
 \setlength{\parsep}{0pt}
 \setlength{\listparindent}{0pt}
 \setlength{\topsep}{0pt}
 }}{\end{list}}

\newcommand{\from}{\colon\!}

\newcommand{\R}{\mathbb{R}}

\newcommand{\B}{B}

\newcommand{\Sc}{\mathcal{S}}

\renewcommand{\vec}[1]{\boldsymbol{#1}}
\renewcommand{\phi}{\varphi}

\newcommand{\KL}{D_{\mathrm{KL}}}
\newcommand{\dom}{\mathrm{dom}}
\newcommand{\Exp}{\mathbf{E}}
\newcommand{\var}{\mathrm{var}}

\newcommand{\cov}{\mathrm{cov}}

\title{Transductive Boltzmann Machines}

\author{
  Mahito Sugiyama \\
  National Institute of Informatics\\
  \texttt{mahito@nii.ac.jp} \\
  \And
  Koji Tsuda\\
  The University of Tokyo\\
  RIKEN AIP; NIMS\\
  \texttt{tsuda@k.u-tokyo.ac.jp}\\
  \And
  Hiroyuki Nakahara\\
  RIKEN Center for Brain Science\\
  \texttt{hiro@brain.riken.jp}
}

\begin{document}

\maketitle

\begin{abstract}
 We present \emph{transductive Boltzmann machines} (TBMs), which firstly achieve transductive learning of the Gibbs distribution.
While exact learning of the Gibbs distribution is impossible by the family of existing Boltzmann machines due to combinatorial explosion of the sample space,
TBMs overcome the problem by adaptively constructing the minimum required sample space from data to avoid unnecessary generalization.
 We theoretically provide \emph{bias-variance decomposition} of the KL divergence in TBMs to analyze its learnability, and empirically demonstrate that TBMs are superior to the fully visible Boltzmann machines and popularly used restricted Boltzmann machines in terms of efficiency and effectiveness.
\end{abstract}

\section{Introduction}
\emph{Transductive learning} is a type of machine learning which directly performs inference on particular given objects without learning unnecessary general rules.
\citet{Vapnik00} describes ``If you are limited to a restricted amount of information, do not solve the particular problem you need by solving a more general problem''.
Since transductive learning is in principle easier than \emph{inductive learning}, which requires to learn general rules for inference on unknown objects, it has drawn considerable attention in the field of machine learning.

However, transductive learning has been largely ignored in learning of the \emph{Gibbs distribution}, which is a fundamental problem for \emph{energy-based models}~\cite{LeCun07}.
Although various types of Boltzmann machines have been proposed for the task, including restricted Boltzmann machines (RBMs)~\cite{Smolensky86,Hinton02} and deep Boltzmann machines (DBMs)~\cite{Salakhutdinov09,Salakhutdinov12}, none of them allows exact learning.
This is why the sample space of the Gibbs distribution is always fixed to $\{0, 1\}^{|V|}$, or the power set $2^V$, of the set $V$ of variables to infer probability of any outcome for induction, which causes combinatorial explosion of the sample space.

Nevertheless, inference on the entire space $\{0, 1\}^{|V|}$ is not always essential.
The motivating application is neural activity analysis~\cite{Ganmor11,Ioffe17,Koster14,Watanabe13,Yu11},
where the goal is to investigate (potentially higher-order) interactions among neurons (variables) by fitting energy-based models.
Since prediction on unknown neural activity is not particularly interesting, generalization from a particular dataset to the entire space $\{0, 1\}^{|V|}\!$ is optional.
Therefore transductive learning to perform inference on only data should be desirable.
The transductive approach has been implicitly employed by~\citet{Ganmor11}.

Here we propose \emph{transductive Boltzmann machines} (TBMs), which adaptively construct the minimum required sample space $S \subset 2^V$ of the Gibbs distribution from data and realize transductive learning \emph{from particular to particular} without solving the unnecessary general task.
Advantages are:
\begin{enumerate}
 \item Learning is efficient. The time complexity of computing the gradient of a parameter is independent of the number of variables and linear in the sample size.
 \item Learning is exact. None of approximation method such as Gibbs sampling is required.
 \item Learning is optimal. Convergence to the global optimum that maximizes the (log-)likelihood is always guaranteed.
\end{enumerate}
Moreover, we theoretically analyze TBMs and achieve \emph{bias-variance decomposition} of the Kullback--Leibler (KL) divergence using \emph{information geometry}~\cite{Amari16}.
In particular, our method can be viewed as generalization of the information geometric hierarchical log-linear model introduced by~\citet{Amari01,Nakahara02}.

This paper is organized as follows.
Section~\ref{sec:tbm} introduces the transductive Boltzmann machines (TBMs).
Section~\ref{subsec:formulation} formulates the TBMs, Section~\ref{subsec:learning} gives a learning algorithm, Section~\ref{subsec:param} discusses how to choose parameters, and Section~\ref{subsec:biasvariance} provides bias-variance decomposition.
We empirically examine TBMs in Section~\ref{sec:exp} and conclude the paper in Section~\ref{sec:conclusion}.

\section{Transductive Boltzmann Machines}\label{sec:tbm}
To introduce our proposal of \emph{transductive Boltzmann machines} (TBMs), we first prepare basic notion of Boltzmann machines.
Let $V$ be the set of (visible) variables.
Given a \emph{Boltzmann machine}~\cite{Ackley85}, which is represented as a graph $(V, E)$ with the vertex set $V$ and the edge set $E$, the \emph{energy} for a variable configuration $\vec{x} \in \{0, 1\}^{|V|}$ is given by
\begin{align*}
 E(\vec{x}; \vec{b}, \vec{w}) = - \sum_{i \in V} b_i x_i - \sum_{i, j \in E} w_{ij} x_i x_j,
\end{align*}
where each $x_i$ is the binary state of an variable $i \in V$, $\vec{b}$ and $\vec{w}$ are called bias and weight, respectively.
The \emph{Gibbs distribution} for the sample space $\Omega = \{0, 1\}^{|V|}\!$ with the \emph{partition function} $Z$ is defined as
\begin{align*}
 p(\vec{x}; \vec{b}, \vec{w}) = \frac{1}{Z} \exp\bigl(-E(\vec{x}; \vec{v}, \vec{w})\bigr),\quad
 Z = \sum_{\mathclap{\vec{x} \in \Omega}} \exp\bigl(-E(\vec{x}; \vec{v}, \vec{w})\bigr),
\end{align*}

We introduce alternative set-theoretic notation of Boltzmann machines.
We represent each binary vector $\vec{x} \in \{0, 1\}^{|V|}$ as the set of indices of ``1'', that is, $\{i \in V \mid x_i = 1\}$.
The sample space $\Omega = \{0, 1\}^{|V|}$ becomes the power set $2^V$ of $V$.
We represent bias and weight by a function $\theta \from B \to \R$, where the domain $B = \{x \in 2^V \mid |x| = 1$ or $x \in E\}$ for a Boltzmann machine $(V, E)$.
For each $x \in 2^V$, the energy is re-written as
\begin{align}
 \label{eq:energy}
 E(x; \theta) &= - \sum_{s \in B} \zeta(s, x)\theta(s),\quad
 \zeta(s, x) = \left\{
 \begin{array}{ll}
   1 & \text{if } s \subseteq x,\\
   0 & \text{otherwise}.
 \end{array}
  \right.
\end{align}
For the partition function $Z$, we denote by $\psi(\theta) = \log Z$. Hence each probability of the resulting Gibbs distribution $P$ can be written as
\begin{align}
 \label{eq:prob}
 \log p(x; \theta) = -E(x; \theta) - \psi(\theta) = \sum_{s \in B} \zeta(s, x)\theta(s) - \psi(\theta).
\end{align}
In what follows, we use the symbol $\bot$ to represent the empty set $\emptyset$. Note that we have
\begin{align*}
 \log p\bigl((0, 0, \dots, 0)\bigr) = \log p(\bot) = -\psi(\theta) = -\log Z.
\end{align*}
We use uppercase letters $P, Q, R, \dots$ for Gibbs distributions and lowercase letters $p, q, r, \dots$ for the corresponding probability.

\subsection{Formulation}\label{subsec:formulation}
We present TBMs, which adaptively construct the sample space $S \subseteq \Omega = 2^V$ of the Gibbs distribution from a given dataset $D \subseteq 2^V$.
The dataset $D$ is naturally assumed to be a multiset, that is, multiple instances of elements are allowed.

Let $V$ be the set of variables, 
Given the domain of parameters $B \subseteq 2^V$ and a dataset $D \subseteq 2^V$.
The sample space $S$ of the Gibbs distribution $P$ generated by a \emph{transductive Boltzmann machine} (TBM) is defined as
\begin{align}
 \label{eq:domTBM}
 S = B \cup D \cup \{\bot\},
\end{align}
where $D$ is treated as a unique set in the union.
The energy and probability for all $x \in S$ in a TBM is defined by Equation~\eqref{eq:energy} and~\eqref{eq:prob}, respectively, and the (logarithm of) partition function is given as
\begin{align}
 \label{eq:psi}
 \psi(\theta) = \log \sum_{x \in S} \exp\bigl(-E(x; \theta)\bigr)
\end{align}
for the parameter function $\theta \from B \to \R$, where $\psi(\theta) = -p(\bot)$ holds.

Note that $B \not= D$ in general.
Since the cardinality of each element $x \in B$ coincides with the \emph{order} of interactions, TBMs can treat higher-order interactions among variables if $|x| > 2$ for some $x \in B$, which have been considered by \citet{Sejnowski86,Min14}.
The parameter domain $B$ defines the model of TBMs, thus the problem of how to choose $B$ corresponds to how to determine the graph structure $(V, E)$ in Boltzmann machines, which belongs to the model selection problem.
We will discuss how to determine the parameter domain $B$ in Section~\ref{subsec:param}.

\subsection{Learning}\label{subsec:learning}
Learning of a TBM is achieved by maximizing the log-likelihood or, equivalently, minimizing the KL divergence, which is the same criterion with the existing Boltzmann machines.
We introduce $\eta \from S \to \R$ corresponding to the \emph{expectation}, defined as
\begin{align}
 \label{eq:eta}
 \eta(x) = \sum_{s \in S} \zeta(x, s) p(x).
\end{align}
It is clear that $\eta(\{i\}) = \Exp[x_i] = \Pr(x_i = 1)$, $\eta(\{i, j\}) = \Exp[x_i x_j] = \Pr(x_i = x_j = 1)$, $\dots$ for $i, j \in V$.
For a dataset $D$, we use the empirical distribution $\hat{P}$ defined as $\hat{p}(x) = \mathbf{1}_D(x) / |D|$, where $\mathbf{1}_D(x)$ is the multiplicity function denoting how many times $x$ appears in the multiset $D$. It follows that $|D| = \sum_{x \in D} \mathbf{1}_D(x)$. We denote by $N = |D|$ and $n = |V|$.

First we show that the log-likelihood $L_D(P)$ of a Gibbs distribution $P$ is \emph{concave with respect to} $\theta(x)$ for any $x \in B$.
This is from
\begin{align*}
 L_D(P) = \sum_{x \in D} \log p(x; \theta)
  = \sum_{x \in D} \sum_{s \in B} \zeta(s, x) \theta(s) - N\psi(\theta)
\end{align*}
and $\psi(\theta)$ is convex with respect to any $\theta(x)$ from Equation~\eqref{eq:psi}.

Next we prove that the log-likelihood $L_D(P)$ is maximized if and only if
\begin{align}
 \label{eq:condition}
 \eta(x) = \hat{\eta}(x),\quad \forall x \in B,
\end{align}
where $\hat{\eta}(x)$ is obtained in Equation~\eqref{eq:eta} by replacing $p$ with $\hat{p}$, which corresponds to the expectation of the empirical distribution.
The gradient of $L_D(P)$ with respect to $\theta(y)$ with $y \in B$ is obtained as
\begin{align*}
 \frac{\partial}{\partial \theta(y)} L_D(P)
 &= \frac{\partial}{\partial \theta(y)} \sum_{x \in D} \log p(x; \theta)
  = \frac{\partial}{\partial \theta(y)} \sum_{x \in D} \sum_{s \in B} \zeta(s, x) \theta(s) - \frac{\partial}{\partial \theta(y)} N\psi(\theta)\\
 &= \sum_{x \in D} \zeta(y, x) - N\frac{\partial \psi(\theta)}{\partial \theta(y)}.
\end{align*}
We have $\sum_{x \in D} \zeta(y, x) = N\hat{\eta}(y)$ as $D$ is a multiset and
\begin{align*}
 \frac{\partial \psi(\theta)}{\partial \theta(y)}
 &= \frac{\partial}{\partial \theta(y)} \log \sum_{x \in S} \exp\bigl(-E(x; \theta)\bigr)
  = \frac{1}{\exp(\psi(\theta))} \frac{\partial}{\partial \theta(y)} \sum_{x \in S} \exp\bigl(-E(x; \theta)\bigr)\\
 &= \frac{1}{\exp(\psi(\theta))} \sum_{x \in S} \exp\bigl(-E(x; \theta)\bigr) \frac{\partial}{\partial \theta(y)} \bigl(-E(x; \theta)\bigr)\\
 &= \frac{1}{\exp(\psi(\theta))} \sum_{x \in S} \exp\bigl(-E(x; \theta)\bigr) \zeta(y, x) = \sum_{x \in S} \zeta(y, x) p(x) = \eta(y).
\end{align*}
Hence it follows that
\begin{align*}
 \frac{\partial}{\partial \theta(y)} L_D(P) = N \bigl(\, \hat{\eta}(y) - \eta(y) \,\bigr).
\end{align*}
Moreover, the KL divergence $\KL(\hat{P}, P) = \sum_{x \in S}\hat{p}(x) \log(\hat{p}(x) / p(x)) = - (1 / N)L_D(P) - H(\hat{P})$,
where $H(\hat{P})$ is the entropy of $\hat{P}$ and independent of $\theta$.
Therefore, together with the concavity of $L_D(P)$, the Gibbs distribution $P$ simultaneously maximizes the log-likelihood $L_D(P)$ and minimizes the KL divergence $\KL(\hat{P}, P)$ if and only if Equation~\eqref{eq:condition} is satisfied.

Our result means that TBMs can be optimized by the gradient ascent strategy.
We show a gradient ascent algorithm for learning of TBMs in Algorithm~\ref{alg:gradTBM}, where $\varepsilon$ is a learning rate.
The time complexity of each iteration for updating $\theta(x)$ is $O(|D| + |B|)$, hence the total time complexity is $O(h|B||D| + h|B|^2)$, where $h$ is the number of iteration.
In the standard Boltzmann machines, the sample space $\Omega = 2^V$, which means that the time complexity for computing $\eta(x)$ to get the gradient of the parameter $\theta(x)$ (bias or weight) is $O(2^{|V|})$ and exact computation is infeasible due to the combinatorial explosion.
Moreover, computing the partition function involves the sum of energy across the entire space $2^V$, which again causes the combinatorial explosion.
In contrast, the complexity of TBMs is independent of the number $|V|$ of variables and linear in the sample size $N = |D|$.
Hence efficient and exact computation is achieved.

\IncMargin{12pt}
\begin{algorithm}[t]
 \begin{small}
 \SetKwInOut{Input}{input}
 \SetKwInOut{Output}{output}
 \SetFuncSty{textrm}
 \SetCommentSty{textrm}
 \SetKwFunction{GradientAscent}{{\scshape GradientAscent}}
 \SetKwProg{myfunc}{}{}{}

 \myfunc{\GradientAscent{$D$, $B$}}{
 Compute $\hat{\eta}$ from $D$ using Equation~\eqref{eq:eta};\\
 Initialize $\theta$, $p$, and $\eta$; \tcp*[f]{e.g.\ $\theta(x) = 0$ and $p(x) = 1/|S|$ for all $x \in B$}\\
 \Repeat{convergence of $\theta$\label{line:endloop}}{\label{line:startloop}
 \ForEach{$x \in B$}{
 $\mu \gets \varepsilon (\hat{\eta}(x) - \eta(x))$;\\
 $\theta(x) \gets \theta(x) + \mu$; \tcp*[f]{update $\theta$}\\
 \lForEach(\tcp*[f]{update probability}){$s \supseteq x$}{$p(s) \gets p(s) \exp(\mu)$}
 }
 $\psi(\theta) \gets \log \sum_{x \in S} p(x)$; \tcp*[f]{update partition function}\\
 \lForEach(\tcp*[f]{normalize probability}){$x \in S$\,}{$p(x) \gets p(x) \exp(-\psi(\theta))$}
 \lForEach(\tcp*[f]{update $\eta$}){$x \in B$}{$\eta(x) \gets \sum_{s \supseteq x} p(s)$}
 }
 Output $\theta$;
 }
 \caption{Learning of TBM by gradient ascent.}
 \label{alg:gradTBM}
 \end{small}
\end{algorithm}
\DecMargin{12pt}

\subsection{Parameter Selection}\label{subsec:param}
We provide a guideline of how to choose the domain $B \subseteq 2^V$ of parameters, which can be selected by the user, although it is hard to find the optimal $B$ without the specific domain knowledge of variable connections.
This is the same problem with how to determine the network structure in neural networks.

We propose to set $B$ as variable combinations that occur \emph{frequently enough} in a dataset, that is, $\hat{\eta}(x) \ge \sigma$ for $x \in B$ with some threshold $\sigma$.
This is why $\hat{\eta}$ is used as the condition of the optimality in Equation~\eqref{eq:condition} and small $\hat{\eta}$ implies that it might not contribute to the resulting Gibbs distribution.
In addition, it is important to restrict higher-order interactions to examine their contribution.
Hence the parameter domain $B$ is formulated as
\begin{align}
 \label{eq:adaptiveB}
 B = \Set{x \in 2^V | \hat{\eta}(x) \ge \sigma \text{ and } |x| \le k},
\end{align}
and the threshold $\sigma \in [0, 1]$ and the upper bound $k$ of the order are the input parameters to TBMs.
Note that $\sigma = 0$ is possible, where all elements $x$ such that $|x| \le k$ are added to $B$.

Interestingly, \emph{frequent itemset mining}~\cite{Agrawal94,Aggarwal14FPM} studied in data mining can exactly solve this problem of finding $B$.
This technique offers to enumerate all frequent itemsets, where an \emph{itemset} $x$ is an element of $2^V$ and it is said to be frequent if $\hat{\eta}(x) \ge \sigma$.
Moreover, we can include the cardinality constraint $k$ on each itemset in the enumeration process.
This means that the set of frequent itemsets coincides with $B$ introduced in Equation~\eqref{eq:adaptiveB}.
Thus we can simply apply a frequent itemset mining algorithm, e.g., LCM~\cite{Uno04} known to be the fastest mining algorithm, to efficiently obtain the parameter domain $B$ from a dataset $D$.
It is interesting to note that itemset mining has been also used in neural activity analysis~\cite{Gruen13} and genome wide association studies~\cite{Zhang14plos}.
We summarize the overall algorithm of TBMs in Algorithm~\ref{alg:TBM}.
Since frequent itemset mining will find massive number of itemsets if the threshold $\sigma$ is small, we recommend to start trying relatively high $\sigma$ in TBMs or small $k$ and decrease $\sigma$ or increase $k$ if $B$ is empty or too small.

In the Gibbs distribution $P$ represented by a TBM, the probability $p(x) > 0$ must be satisfied for all $x \in S$, otherwise $\theta$ diverges.
Unfortunately, there exists some $B$ which causes this situation when the learning condition in Equation~\eqref{eq:condition} is satisfied.
For example, let $V = \{1, 2\}$, $S = \{\emptyset, \{1\}, \{2\}, \{1, 2\}\}$, $B = \{\{1\}, \{1, 2\}\}$, and $\hat{\eta}(\{1\}) = \hat{\eta}(\{1, 2\}) = 0.4$.
Then Equation~\eqref{eq:condition} is satisfied only if $p(\{1\}) = 0$ and $p(\{1, 2\}) = 0.4$.
This problem is solved if we remove $\{1\}$ or $\{1, 2\}$ from $B$.
Hence, in learning of TBMs in the gradient ascent algorithm in Algorithm~\ref{alg:gradTBM}, one needs to monitor the behavior of $\theta$ and remove the corresponding element from $B$ when it starts diverging.
Note that this problem exists not only in TBMs but in the standard Boltzmann machines.
But it has not been investigated before.

\IncMargin{10pt}
\begin{algorithm}[t]
 \begin{small}
 \SetKwInOut{Input}{input}
 \SetKwInOut{Output}{output}
 \SetFuncSty{textrm}
 \SetCommentSty{textrm}
 \SetKwFunction{TBM}{{\scshape TBM}}
 \SetKwFunction{GradientAscent}{{\scshape GradientAscent}}
 \SetKwProg{myfunc}{}{}{}

 \myfunc{\TBM{$D$, $\sigma$, $k$}}{
 $B \gets \{x \in 2^V \mid \hat{\eta}(x) \ge \sigma \text{ and } |x| \le k\}$; \tcp*[f]{frequent itemset mining algorithm can be used}\\
 $\theta \gets$ \GradientAscent{$D$, $B$}; \tcp*[f]{see Algorithm~\ref{alg:gradTBM}}\\
 Output $(\theta, B)$;
 }
 \caption{TBM.}
 \label{alg:TBM}
\end{small}
\end{algorithm}
\DecMargin{10pt}

\subsection{Bias-Variance Decomposition}\label{subsec:biasvariance}
To theoretically analyze learnability of TBMs, we provide bias-variance decomposition of the KL divergence, which is a fundamental analysis of machine learning methods.
Since the sample space $S$ is a subset of $2^V$, it is always a \emph{partially ordered set} (poset)~\cite{Gierz03} with respect to the inclusion relationship ``$\subseteq$''.
This directly leads to the following result:
\begin{proposition}
 \label{prop:loglinear}
 TBMs belong to the log-linear model on posets introduced by~\citet{Sugiyama17ICML}.
 Probability $p$ and the parameter $\theta$ in Equation~\eqref{eq:prob} and $\eta$ in Equation~\eqref{eq:eta} correspond to those in Equations~\textup{(8)} and~\textup{(7)} in~\cite{Sugiyama17ICML}, respectively.
\end{proposition}

The log-linear model on posets is an extension of the hierarchical model of probability distributions studied in information geometry~\cite{Amari01,Nakahara02}, and the following holds from Proposition~\ref{prop:loglinear}.
The set of Gibbs distributions $\vec{\Sc}$ with the sample space $S$ is always a \emph{dually flat manifold}~\cite{Amari09}, and the pair of functions $(\theta, \eta)$ is a dual coordinate system of the manifold (Theorem~2 in~\cite{Sugiyama17ICML}), which is connected with the Legendre transformation:
\begin{align*}
 \theta(x) = \frac{\partial \phi(\eta)}{\partial \eta(x)}\quad\text{and}\quad
 \eta(x) = \frac{\partial \psi(\theta)}{\partial \theta(x)},\quad
 \forall x \in S^+,
\end{align*}
where $\phi(\eta) = \sum_{x \in S} p(x) \log p(x)$ and $S^+ = S \setminus \{\bot\}$.

The dually flat structure of $\vec{\Sc}$ gives us \emph{Pythagorean theorem}.
Let us consider two submanifolds:
\begin{align*}
 \vec{\Sc}_{\alpha} &= \Set{P \in \vec{\Sc} | \theta(x) = \alpha(x) \text{ for all } x \in \dom(\alpha)},\\
 \vec{\Sc}_{\beta} &= \Set{P \in \vec{\Sc} | \eta(x) = \beta(x) \text{ for all } x \in \dom(\beta)}
\end{align*}
specified by two functions $\alpha, \beta$ with $\dom(\alpha), \dom(\beta) \subseteq S^+$, where the former submanifold $\vec{\Sc}_{\alpha}$ has constraints on $\theta$ while the latter $\vec{\Sc}_{\beta}$ has those on $\eta$.
Submanifolds $\vec{\Sc}_{\alpha}$ and $\vec{\Sc}_{\beta}$ are called $e$-flat and $m$-flat, respectively~\cite[Chapter~2.4]{Amari16}.
Assume that $\dom(\alpha) \cup \dom(\beta) = S^+$ and $\dom(\alpha) \cap \dom(\beta) = \emptyset$.
Then the intersection $\vec{\Sc}_{\alpha} \cap \vec{\Sc}_{\beta}$ is always a singleton, that is, the distribution $Q$ satisfying $Q \in \vec{\Sc}_{\alpha}$ and $Q \in \vec{\Sc}_{\beta}$ always uniquely exists, leading to the Pythagorean theorem:
\begin{align}
 \label{eq:Pythagorean}
 \KL(P, R) = \KL(P, Q) + \KL(Q, R),\quad
 \KL(R, P) = \KL(R, Q) + \KL(Q, P)
\end{align}
for any $P \in \vec{\Sc}_{\alpha}$ and $R \in \vec{\Sc}_{\beta}$.

Using Pythagorean theorem, we achieve bias-variance decomposition in learning of TBMs.
Our idea is to decompose the expectation of the KL divergence $\Exp[\KL(P^*$, $\hat{P}_{\B})]$ from the true (unknown) Gibbs distribution $P^*$ to the MLE (maximum likelihood estimation) $\hat{P}_{\B}$ of an empirical distribution $\hat{P}$ learned by a TBM with a fixed parameter domain $B$, and decompose it using the information geometric property.
Interestingly, we can simply obtain the lower bound of the variance as $|B| / 2N$ using only the number of parameters $|B|$ and the sample size $N = |D|$, which is \emph{independent of the sample space} $S$.
This theorem also applies to the fully visible Boltzmann machines with $S = 2^V$.

\begin{theorem}[Bias-variance decomposition of the KL divergence]\label{theorem:biasvariance}
 Given a parameter domain $B$ and sample space $S$ such that $B \subseteq S \subseteq 2^V$.
 Let $P^* \in \vec{\Sc}$ be the true Gibbs distribution, $P^*_{\B}, \hat{P}_{\B} \in \vec{\Sc}(B)$ be the MLEs of $P^*$ and an empirical distribution $\hat{P}$ learned by a TBM, respectively.
 We have
 \begin{align*}
  \Exp\Bigl[\KL(P^*, \hat{P}_{\B})\Bigr]
  = \KL(P^*, P^*_{\B}) + \Exp\Bigl[\KL(P^*_{\B}, \hat{P}_{\B})\Bigr]
  = \underbrace{\KL(P^*, P^*_{\B})}_{\textup{bias}} + \underbrace{\var(P^*_{\B})}_{\textup{variance}},
 \end{align*}
 where the variance is given as
\begin{align*}
 \var(P^*_{\B}) = \Exp\left[\psi(\hat{\theta}_{\B})\right] - \psi(\theta^*_{\B}) \simeq \frac{1}{2}\sum_{s \in B}\sum_{u \in B} \cov\!\left(\hat{\theta}_B(s), \hat{\theta}_B(u)\right) g_{su}
 \ge \frac{|B|}{2N} + O(N^{-1.5})
\end{align*}
with the equality holding when the sample size $N \to \infty$, where $\cov(\hat{\theta}_{\B}(s), \hat{\theta}_{\B}(u))$ denotes the error covariance between $\hat{\theta}_{\B}(s)$ and $\hat{\theta}_{\B}(u)$ and $g_{su} = \partial \eta^*_B(s) / \partial \theta^*_B(u)$ is the Fisher information.
\end{theorem}

\textit{Proof}.\hspace*{.5em}
For two submanifolds:
\begin{align*}
 \vec{\Sc}(B) = \Set{P \in \vec{\Sc} | \theta(x) = 0 \text{ for all } x \not\in B},\quad\!
 \vec{\Sc}(P^*) = \Set{P \in \vec{\Sc} | \eta(x) = \eta^*(x) \text{ for all } x \in B}
\end{align*}
with $P^* \in \vec{\Sc}(P^*)$ and $\hat{P}_B \in \vec{\Sc}(B)$, we apply Pythagorean theorem, yielding
\begin{align*}
 \Exp\Bigl[\KL(P^*, \hat{P}_{\B})\Bigr]
 = \Exp\Bigl[\KL(P^*, P^*_{\B}) + \KL(P^*_{\B}, \hat{P}_{\B})\Bigr]
 = \KL(P^*, P^*_{\B}) + \Exp\Bigl[\KL(P^*_{\B}, \hat{P}_{\B})\Bigr],
\end{align*}
where $\vec{\Sc}(P^*) \cap \vec{\Sc}(B) = \{P^*_{\B}\}$.
The second term is
\begin{align*}
 \Exp\Bigl[\KL(P^*_{\B}, \hat{P}_{\B})\Bigr]
 &= \Exp\left[\sum\nolimits_{x \in S} p^*_{\B}(x) \log (p^*_{\B}(x) / \hat{p}_{\B}(x))\right]\\
 &= \Exp\left[\sum\nolimits_{x \in S} p^*_{\B}(x)\! \left(\sum\nolimits_{s \in B} \zeta(s, x)\! \left(\theta^*_{\B}(s) - \hat{\theta}_{\B}(s)\right) - \left(\psi(\theta^*_{\B}) - \psi(\hat{\theta}_{\B})\right)\right)\right]\\
 &= \Exp\left[\sum\nolimits_{s \in B} \eta^*_{\B}(s)\left(\theta^*_{\B}(s) - \hat{\theta}_{\B}(s)\right)\right] + \Exp\left[\psi(\hat{\theta}_{\B}) - \psi(\theta^*_{\B})\right]\\
 &= \Exp\left[\psi(\hat{\theta}_{\B})\right] - \psi(\theta^*_{\B}),
\end{align*}
Since $\psi(\theta)$ is the function of $\theta(x)$, $x \in B$, we use the second-order approximation of the Taylor series expansion~\cite[Section~4.3.2]{Ang06}, which is given for a function $y = g(x_1, \dots, x_n)$ as
\begin{align*}
 y = g(a_1, a_2, \dots, a_n) + \sum_{i = 1}^n(x_i - a_i)\frac{\partial g}{\partial x_i} + \frac{1}{2}\sum_{i = 1}^n\sum_{j = 1}^n(x_i - a_i)(x_j - a_j)\frac{\partial^2 g}{\partial x_i \partial x_j}.
\end{align*}
Thus $\Exp[\KL(P^*_{\B}, \hat{P}_{\B})] = \Exp[\psi(\hat{\theta}_{\B})] - \psi(\theta^*_{\B})$ is approximated as
\begin{align*}
 \Exp\Bigl[\KL(P^*_{\B}, \hat{P}_{\B})\Bigr]
 &\simeq \psi(\theta^*_{\B}) + \frac{1}{2}\sum_{s \in B}\sum_{u \in B}\cov\!\left(\hat{\theta}_{B}(s), \hat{\theta}_{B}(u)\right)\frac{\partial^2 \psi(\theta^*_B)}{\partial \theta^*_B(s) \partial \theta^*_B(u)} - \psi(\theta^*_{\B})\\
 &= \frac{1}{2}\sum_{s \in B}\sum_{u \in B}\cov\!\left(\hat{\theta}_{B}(s), \hat{\theta}_{B}(u)\right) g_{su}.
\end{align*}
We can use the Cram\'er--Rao bound~\cite[Theorem 7.1]{Amari16} of the covariance matrix since $\theta_{\B}(x)$ for any $x \in B$ is unbiased, where we additionally have the term $O(N^{-1.5})$ as we used the second-order Taylor approximation of the expectation.
Finally we have
\begin{align*}
 \Exp\Bigl[\KL(P^*_{\B}, \hat{P}_{\B})\Bigr]
 &\ge \frac{1}{2N}\sum_{s \in B}\sum_{u \in B} (g^{-1})_{su} g_{su} + O(N^{-1.5}) = \frac{|B|}{2N} + O(N^{-1.5})
\end{align*}
with the equality holding when $N \to \infty$.$\hfill\square$

Since $\KL(P^*, P^*_{B_1}) \le \KL(P^*, P^*_{B_2})$ always holds if $B_1 \supseteq B_2$ as $\vec{\Sc}(B_1) \supseteq \vec{\Sc}(B_2)$, our result supports intuitive property of typical machine learning algorithms: If we extend the parameters $B$, the bias decreases while the variance increases.

\begin{figure}[t]
 \centering
 \includegraphics[width=\linewidth]{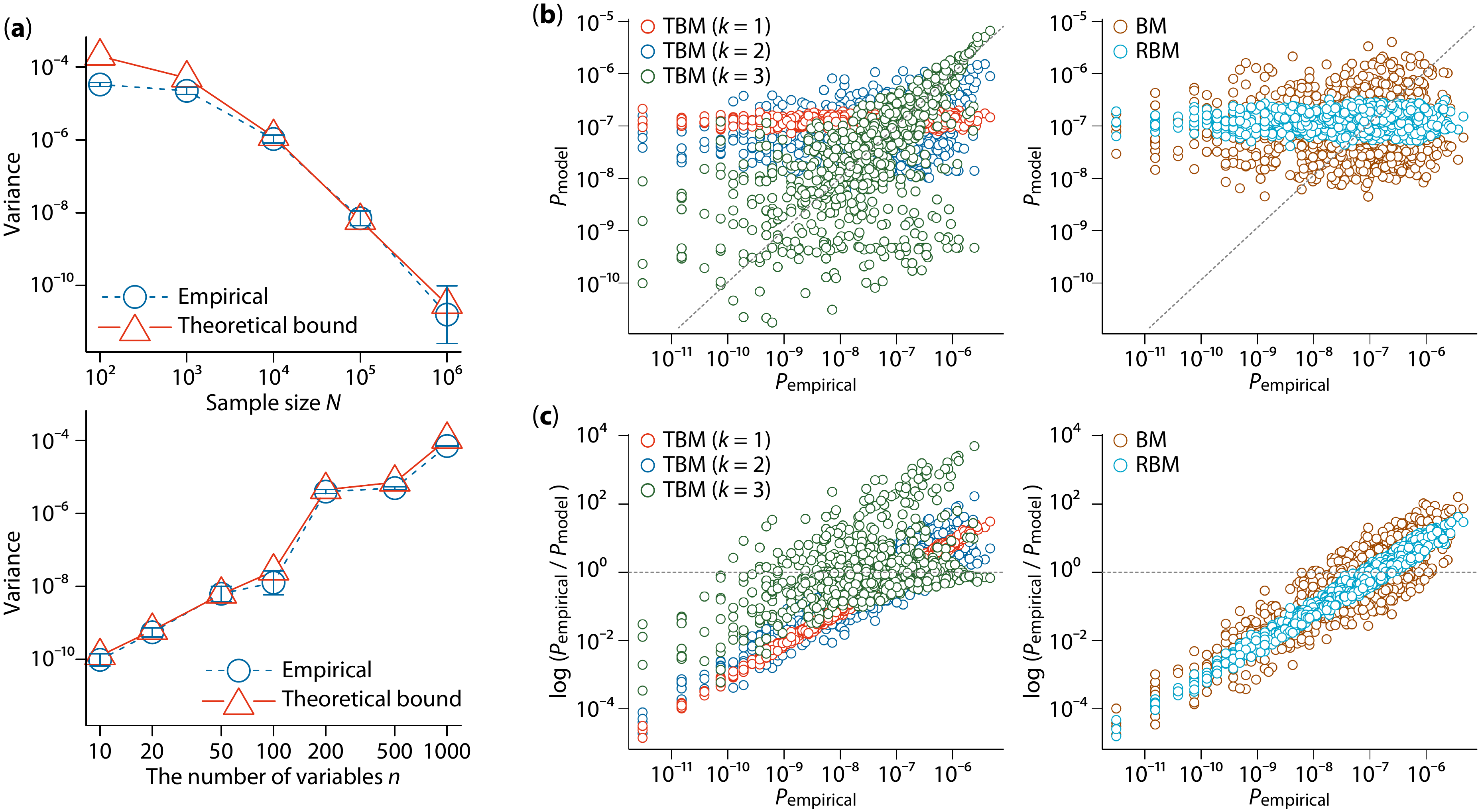}
 \caption{(\textbf{a}) Empirical evaluation of variance. Empirically estimated variances (blue, dotted lines) and theoretically obtained lower bounds (red, solid line) for $n =$ 50 (top) and $N =$ 100,000 (bottom). (\textbf{b}) The learned model distribution $P$ against the empirical distribution $\hat{P}$ for TBMs with $k = 1, 2, 3$ (left) and BM, RBM (right). (\textbf{c}) The log-likelihood $\log (\hat{P} / P)$ against the empirical distribution $\hat{P}$ for TBMs with $k = 1, 2, 3$ (left) and BM, RBM (right).}
 \label{figure:synth}
\end{figure}

We empirically demonstrate the tightness of the lower bound of the variance in this theorem, as shown in Figure~\ref{figure:synth}(\textbf{a}).
To obtain the variance $\var(P^*_{\B}) = \Exp[\KL(P^*_{\B}, \hat{P}_{\B})]$, first we fix a true distribution $P^*$ generated from the uniform distribution with its sample space $S$ with $|S| = 1,000$ and get $P^*_{\B}$ learned by a TBM with $\sigma = 0.37$ and $k = 2$, which gives a reasonable amount of parameters $B$.
Then the lower bound is obtained as $|B| / 2N$.
In each trial, we repeat $100$ times generating a sample $D$ with the size $N$ from $P^*$ and learned $\hat{P}_{B}$ with fixing $S$ and $B$ to directly estimate the variance ($\pm$ its standard deviation).
In Figure~\ref{figure:synth}(\textbf{a}, top) the sample size $N$ is varied from 100 to 1,000,000 with fixing the number of variables $n = |V| = \text{50}$ while in Figure~\ref{figure:synth}(\textbf{a}, bottom) $n$ is varied from 10 to 1,000 with fixing $N = |D| = \text{100,000}$.
These plots clearly show that our lower bound is tight enough across all settings.
The lower bound exceeds the actual variance in some cases, which is due to the approximation error of the Taylor series expansion or fluctuation of random sampling.

\setlength{\textfloatsep}{15pt}

\section{Experiments}\label{sec:exp}
In our simulations, we empirically found the effectiveness and efficiency of transductive Boltzmann machines (TBMs) compared to two representative energy-based models: fully visible Boltzmann machines (BMs) and restricted Boltzmann machines (RBMs), using the setting in the following.
We used Cent OS release 6.9 and ran all experiments on 2.20~GHz Intel Xeon CPU E7-8880 v4 and 3~TB of memory.
All methods, TBMs, BM, and RBMs, were implemented in \texttt{C++} and compiled with \texttt{icpc} 18.0.1.
We used persistent contrastive divergence with a single step of alternating Gibbs sampling (persistent CD-1)~\cite{Hinton02,Tieleman08} in learning of BMs and RBMs.
We measured the effectiveness of respective methods by the reconstruction error (smaller is better), which is defined as the KL divergence $\KL(\hat{P}, P)$ from the empirical distribution $\hat{P}$ of a given dataset $D$ to the learned model distribution $P$.
Since it is impossible to compute the exact partition function in BMs and TBMs and an additional approximation method such as AIS is needed~\cite{Neal01,Salakhutdinov08}, we consistently normalized the learned energy $E(x)$ of each $x \in D$ by the proxy $Z' = \sum_{x \in D} \exp(-E(x))$ in all the three methods to exclude the approximation error of the partition function and realize fair comparison.
Throughout the experiments, the number of iterations for TBMs were up to $10^4$ and those for BMs and RBMs are $10^6$ to ensure the convergence.
Running time in TBMs includes the itemset mining process by LCM~\cite{Uno04} to obtain $B$.

\paragraph{Results on Synthetic Data.}
First we demonstrate the performance of the thee methods on synthetic data and show that TBMs can accurately learn Gibbs distributions without overfitting.
We have randomly generated a dataset $D$, where we first randomly chose $D' \subseteq 2^V$ without any multiplicity and sampled $N$ data points from $D'$ with replacement.
We set $N = |D| = 100,000$, $n = |V| = 20$, and $|D'| = 1,000$, and applied TBM with the cardinality upper bound $k = 1, 2$, or $3$ on each parameter with $\sigma = 0.1$, BM using the same $B$ as the TBM ($k = 2$), and RBM with $100$ hidden variables.
To visualize the performance of these methods, we plot the leaned model distribution $P$ against the empirical distribution $\hat{P}$ in Figure~\ref{figure:synth}(\textbf{b}) and the log-likelihood ratio $\log (\hat{P} / P)$ in Figure~\ref{figure:synth}(\textbf{c}).
The reconstruction errors are 0.29, 0.26, 0.20, 0.41, and 0.30 for TBMs (k = 1, 2, 3), BM, and RBM, respectively.
The results of the TBM with $k = 1$ and the RBM are similar, which does not have second order interactions between visible variables by definition, and those of the TBM with $k = 2$ and the BM using the same parameters $B$ are also similar, which means that TBMs do not overly fit to a random dataset.
Furthermore, we observe that the TBM can achieve more accurate inference if we include higher-order interaction in the TBM with $k = 3$.
\begin{wraptable}{r}{130pt}
 \vspace*{-5pt}
 \setlength{\tabcolsep}{4pt}
 \begin{footnotesize}
 \centering
 \caption{Number of parameters.}\label{table:param}
 \begin{tabular}{p{35pt}rr}
  \toprule
  & TBM\&BM & RBM \\
  \midrule
  20080516\_KO & 305 & 335 \\
  20080624\_WT & 530 & 531 \\
  20080624\_KO & 137 & 194 \\
  20080628\_KO & 899 & 948 \\
  20080628\_WT & 1,865 & 1,891 \\
  20080702\_KO &  45 &  47 \\
  \midrule
  connect & 274 & 1,039 \\
  kosarak & 194 & 82,541 \\
  retail & 128 & 32,941 \\
  chess & 372 & 379 \\
  mushroom &  23 & 239 \\
  pumsb & 3,394 & 6,341 \\
  \bottomrule
 \end{tabular}
  \end{footnotesize}
 \vspace*{-5pt}
\end{wraptable}

\begin{table}[t]
 \setlength{\tabcolsep}{4pt}
 \begin{footnotesize}
  \centering
  \caption{Results on real datasets.}\label{table:real}
   \begin{tabular*}{\textwidth}{p{40pt}>{\raggedleft\arraybackslash}p{35pt}>{\raggedleft\arraybackslash}p{23pt}S[mode=text,tight-spacing,table-parse-only,table-number-alignment=left,text-rm=\bfseries]S[mode=text,tight-spacing,table-parse-only,table-number-alignment=left]S[mode=text,tight-spacing,table-parse-only,table-number-alignment=left]S[mode=text,tight-spacing,table-parse-only,table-number-alignment=left]S[mode=text,tight-spacing,table-parse-only,table-number-alignment=left]S[mode=text,tight-spacing,table-parse-only,table-number-alignment=left]rr}
    \toprule
    Dataset & $N$ & $n$ & \multicolumn{3}{c}{Reconstruction error (KL divergence)} & \multicolumn{3}{c}{Running time (sec.)}\\
    \cmidrule(r){4-6}\cmidrule(r){7-9}
    &&& TBM & BM & RBM & TBM & BM & RBM &\\
    \midrule
    20080516KO & 89,897 &  41 & 7.38E-01 & 2.10E+00 & 1.23E+00 & \Bf{5.01E+00} & 1.35E+01 & 7.41E+00\\
    20080624WT & 53,875 &  37 & 4.64E-02 & 1.63E+00 & 2.20E+00 & \Bf{9.32E+00} & 1.14E+01 & 1.04E+01\\
    20080624KO & 53,558 &  64 & 8.19E-01 & 1.32E+00 & 9.46E-01 & 1.06E+01 & 2.70E+01 & \Bf{7.91E+00}\\
    20080628KO & 71,760 &  72 & 5.19E-02 & 4.36E-01 & 5.63E+00 & \Bf{1.17E+01} & 3.70E+01 & 1.41E+01\\
    20080628WT & 71,862 &  85 & 1.77E-03 & 7.42E+00 & 8.89E+00 & \Bf{7.70E+00} & 4.00E+01 & 2.27E+01\\
    20080702KO & 89,893 &  15 & 1.09E+00 & 2.49E+00 & 2.34E+00 & \Bf{4.90E-01} & 3.91E+00 & 2.56E+00\\
    \midrule
    connect  &  67,557 &   129 & 5.21E-05 & 9.87E-01 & 4.67E+00 & \Bf{5.73E+00} & 9.88E+01 & 2.41E+01 \\
    kosarak  & 990,002 & 41,270 & 1.94E+00 & {---} & 2.00E+00 & \Bf{1.57E+01} & {$\ge$} 1 week & 7.50E+03 \\
    retail   &  88,162 & 16,470 & 1.59E-01 & {---} & 1.53E+00 & \Bf{7.50E-01} & {$\ge$} 1 week & 2.43E+03 \\
    chess    &   3,196 &    75 & 1.01E-02 & 8.96E-02 & 5.55E-01 & \Bf{4.80E-01} & 4.26E+01 & 1.34E+01 \\
    mushroom &   8,124 &   119 & 1.19E-06 & 4.01E-02 & 4.56E-01 & \Bf{1.10E-01} & 8.42E+01 & 1.55E+01 \\
    pumsb    &  49,046 &  2,113 & 1.10E-02 & 1.06E+01 & 3.62E+01 & \Bf{3.47E+01} & 5.56E+04 & 2.50E+02 \\
    \bottomrule
   \end{tabular*}
  \end{footnotesize}
 \end{table}

\vspace*{-10pt}\paragraph{Results on Real Data.}
Next we examine the performance of the three methods on a variety of real datasets.
We collected binary datasets from two domains; neural spiking data~\cite{Zhang14} originally obtained by~\citet{Lefebvre08} and itemset mining benchmark datasets from the FIMI repository\footnote{\url{http://fimi.ua.ac.be/data/}}.
We set $k = 2$ in TBMs and used $\sigma = 0.01$ in both TBMs and BM, hence the parameter domain $B$ is always the same between TBMs and BMs.
The number of parameters in RBMs is $n + n_H + n n_H$ with the number $n_H$ of hidden variables, thereby we set $n_H = \lceil (|B| - n) / (n + 1)\rceil$ to use the same number of parameters with TBMs as much as possible.
The resulting numbers of parameters are shown in Table~\ref{table:param}.
The large difference (e.g.\ in kosarak and retail) cannot be avoided as they are the minimum in RBMs, that is, $2n + 1$ with $n_H = 1$.

Results are shown in Table~\ref{table:real}.
TBMs consistently show the best reconstruction errors (smallest KL divergence) across all datasets and the fastest except for 20080624KO.
Since the time complexity is independent of $n$ in TBMs, it shows two orders of magnitude faster than BMs and RBMs for large $n$ such as in kosarak and retail, while the error of TBMs is smaller than RBMs with limited amount of parameters whose size is two orders of magnitude smaller than those in RBMs.
This is why TBMs avoid unnecessary learning of the Gibbs distribution with the enormous sample space $2^V$ and fit to minimum required space $S$.

\section{Conclusion}\label{sec:conclusion}
This paper firstly shows the powerful potential of transductive learning in energy-based models.
We have proposed to use \emph{transduction} in learning of the Gibbs distribution and presented \emph{transductive Boltzmann machines} (TBMs).
TBMs avoids unnecessary generalization by adaptively constructing the sample space, which leads to efficient and exact learning of the Gibbs distribution,
while exact learning is impossible due to combinatorial explosion of the sample space in the exiting Boltzmann machines.
Our experimental results support the superiority of TBMs in terms of efficiency and effectiveness over the inductive approach by the exiting Boltzmann machines.
This work opens the door of transductive learning to energy-based models.



\end{document}